\documentclass{article}

\usepackage[nonatbib, final]{orl_workshop}
    
\usepackage[utf8]{inputenc} %
\usepackage[T1]{fontenc}    %
\usepackage{bm}
\usepackage{url}            %
\usepackage{booktabs}       %
\usepackage{amsfonts}       %
\usepackage{nicefrac}       %
\usepackage{microtype}      %
\usepackage{todonotes}
\usepackage{graphicx}
\usepackage{enumitem}
\usepackage{tabularx}
\usepackage{xcolor}
\usepackage{hyperref}       %
\usepackage{amsmath}
\usepackage[capitalize]{cleveref}
\usepackage{siunitx}
\usepackage[sort&compress,numbers]{natbib}
\usepackage{array,multirow}
\usepackage{float}
\usepackage{subfig}
\usepackage{colortbl}
\usepackage{authblk}

\usepackage[scaled=0.83]{DejaVuSansMono}

\definecolor{deepmagenta}{rgb}{0.8, 0.0, 0.8}

\title{Learning Object-Centric Video Models \\ by Contrasting Sets}

\newcommand\blfootnote[1]{%
  \begingroup
  \renewcommand\thefootnote{}\footnote{#1}%
  \addtocounter{footnote}{-1}%
  \endgroup
}

\author[2,*]{Sindy Löwe}
\author[1]{Klaus Greff}
\author[1]{Rico Jonschkowski}
\author[1]{Alexey Dosovitskiy}
\author[1]{Thomas Kipf}

\affil[1]{Google Research, Brain Team}
\affil[2]{UvA-Bosch Delta Lab, University of Amsterdam}

\begin{document}
\renewcommand{\t}{_{t}}
\newcommand{\tone}{_{t+1}}
\newcommand{\ttwo}{_{t+2}}
\newcommand{\tthree}{_{t+3}}
\newcommand{\tfour}{_{t+4}}
\newcommand{\avec}{\bm a}
\newcommand{\cvec}{\bm c}
\newcommand{\gvec}{\bm g}
\newcommand{\hvec}{\bm h}
\newcommand{\pvec}{\bm p}
\newcommand{\svec}{\bm s}
\newcommand{\uvec}{\bm u}
\newcommand{\xvec}{\bm x}
\newcommand{\zvec}{\bm z}
\newcommand{\XMat}{\bm X}

\newcommand{\batch}{\mathcal{B}}
\newcommand{\loss}{\mathcal{L}}

\newcommand{\inputt}{\xvec}
\newcommand{\latentrep}{\hvec}
\newcommand{\attn}{\avec}
\newcommand{\attnlogits}{\hat{\attn}}
\newcommand{\update}{\text{update}}
\newcommand{\updates}{\uvec}
\newcommand{\slot}{\svec}
\newcommand{\slots}{\svec}
\newcommand{\slotinit}{\cvec}
\newcommand{\pred}{\pvec}
\newcommand{\projectedslot}{\hat{\pred}}
\newcommand{\globalrep}{\zvec}

\newcommand{\backbone}{f_{\text{enc}}}
\newcommand{\dense}{f_{\text{linear}}}
\newcommand{\mlp}{f_{\text{mlp}}}
\newcommand{\ffw}{f_{\text{ffw}}}
\newcommand{\slotattention}{\text{SlotAttention}}
\newcommand{\transition}{f_{\text{transition}}}
\newcommand{\layernorm}{\text{LayerNorm}}
\newcommand{\posemb}{\text{PositionEmbedding}}
\newcommand{\reshape}{\text{reshape}}
\newcommand{\softmax}{\text{Softmax}}
\newcommand{\deepsets}{\text{DeepSets}}
\newcommand{\weightedmean}{\text{WeightedMean}}
\newcommand{\filtersize}{D_{\text{enc}}}
\newcommand{\slotsize}{D}

\renewcommand{\k}{k}
\newcommand{\q}{q}
\renewcommand{\v}{v}

\newcolumntype{Y}{>{\tiny\arraybackslash}X}

\maketitle
\vspace{-1.6em}
\begin{abstract}
Contrastive, self-supervised learning of object representations recently emerged as an attractive alternative to reconstruction-based training.
Prior approaches focus on contrasting individual object representations (slots) against one another.
However, a fundamental problem with this approach is that the overall contrastive loss is the same for (i) representing a different object in each slot, as it is for (ii) (re-)representing the same object in all slots.
Thus, this objective does not inherently push towards the emergence of object-centric representations in the slots.
We address this problem by introducing a global, set-based contrastive loss: instead of contrasting individual slot representations against one another, we aggregate the representations and contrast the joined sets against one another.
Additionally, we introduce attention-based encoders to this contrastive setup which simplifies training and provides interpretable object masks.
Our results on two synthetic video datasets suggest that this approach compares favorably against previous contrastive methods in terms of reconstruction, future prediction and object separation performance.
\end{abstract}
\vspace{-2em}

\blfootnote{\kern-1.7em$^*$Work done while interning at Google, Contact: \href{mailto:loewe.sindy@gmail.com}{\texttt{loewe.sindy@gmail.com}}}%

\section{Introduction}

Object-centric approaches, for which a scene is represented by a set of object variables (called \textit{slots}), can greatly improve generalization to new situations in an environment or video~\citep{van2018relational,kulkarni2019unsupervised,sun2019stochastic,yi2019clevrer}. Approaches that explicitly model objects can re-use and transfer learned knowledge about the dynamics of individual objects and their interactions, even if the composition of objects in the scene undergoes significant changes.
Recently, contrastive losses have achieved promising results for a variety of image-level tasks~\citep{he2020momentum, chen2020simple,chen2020big}, but have received far less attention for learning object-centric representations. 
All prior approaches to contrastive object discovery \citep{kipf2019contrastive,racah2020slot,huang2020set,anonymous2021systematic} rely on per-slot losses, which have a fundamental flaw: slotwise contrastive losses cannot differentiate between representations in which (i) a different object is represented in each slot or in which (ii) the same object is (re-)represented in all slots. 
As a result, this objective does not inherently enforce diverse slots representations, and needs to rely on additional cues such as object interactions or on explicit regularization.

We expose this issue and propose a solution in terms of a global set contrastive loss (SetCon). Instead of contrasting slot representations directly against one another, we aggregate the set of slots into a global scene representation. This global contrasting approach encourages better coordination between slots, as they are forced to jointly represent the entire scene. 
We further introduce attention-based encoders to this domain using Slot Attention \citep{locatello2020object}, which simplifies training and provides interpretable object masks.
We evaluate our approach on two synthetic video datasets, where we measure performance in terms of the ability of a separately trained decoder to reconstruct either the current time-step or the predicted future time-step, and its ability to separate and locate the different objects present in the scene.

\section{Method}

\begin{figure}[t]
    \centering
    \includegraphics[width=\textwidth]{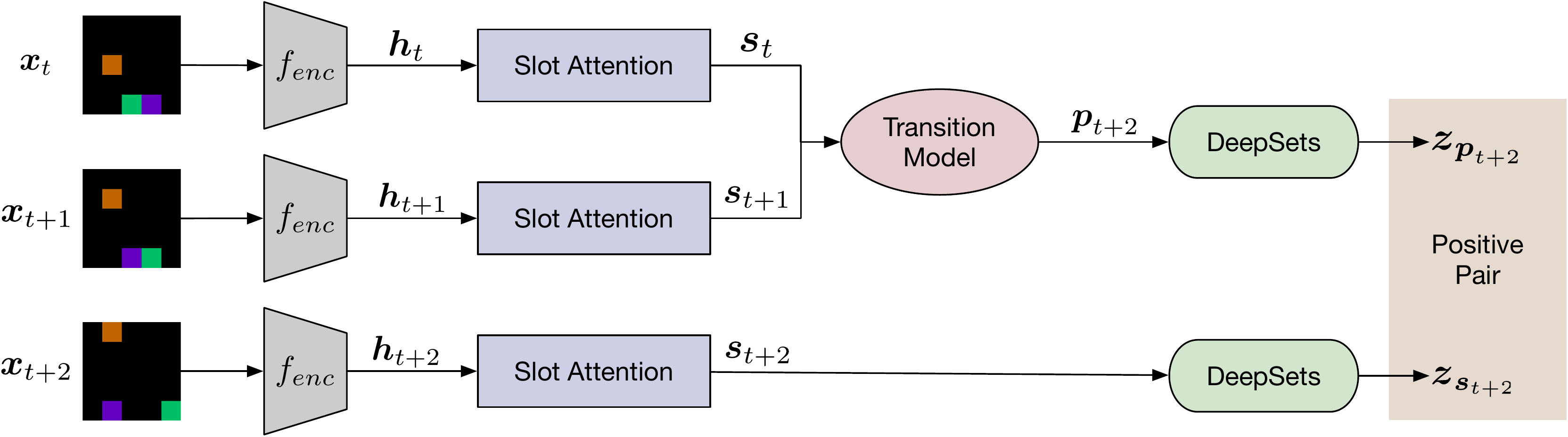}
    \caption{Set Contrastive (SetCon) Slot Attention model.}
    \label{fig:model}
\end{figure}

In this section, we outline the four modules (Backbone, Slot Attention, Transition Model and Set Encoder) that make up the SetCon Slot Attention model (\cref{fig:model}) and describe the proposed Set Contrastive loss for learning object-centric video representations.

\textbf{Backbone} \ We apply an encoding backbone $\backbone$ to an input frame $\xvec\t$ at time step $t$, consisting of a stack of convolutional layers with ReLUs and a linear position embedding:
$\latentrep\t = \backbone(\xvec\t)$.

\textbf{Slot Attention} \ 
The backbone is followed by a Slot Attention module \citep{locatello2020object}, $\slot\t = \slotattention(\latentrep\t)$, applied to hidden representations $\latentrep\t\in\mathbb{R}^{N\times\filtersize}$, where $N$ corresponds to the number of pixels in the (flattened) image, i.e. width$\times$height, and $\filtersize$ is the representation size per pixel. Slot Attention is an iterative attention mechanism that produces $K$ slots $\slot\t = [\slot\t^1, ..., \slot\t^K]$, which can represent individual objects within the input. The two most important steps within each iteration are: (i) slots compete through a softmax attention mechanism that is normalized over the slot dimension and (ii) the final representations are created by aggregating inputs with a weighted mean with the attention matrix $\attn\t$ acting as weights: 
\begin{align}
    \attn\t = \softmax\biggl(\frac{1}{\sqrt{D}} \k(\latentrep\t) \cdot \q(\slotinit)^T\biggr)\in\mathbb{R}^{N\times K} ,\,\,\,
    \updates\t = \weightedmean\biggl(\attn_t, \v(\latentrep\t)\biggr)\in\mathbb{R}^{K\times \slotsize} ,
\end{align}
where $k,q,v$ are linear, learnable projections that map to a common dimension $D$. We  use a small feedforward network $\slots\t = \ffw(\updates\t)$ to arrive at slot representations $\slots\t\in\mathbb{R}^{K\times D}$. 
To simplify training, we do not initialize the slot representations randomly; instead, each slot is initialized with a learned value $\slotinit\in\mathbb{R}^{K\times D}$, and we apply only one iteration of the Slot Attention mechanism.

\textbf{Transition Model} \ 
We apply a transition model with shared parameters to each of the slot representations $\slot\t^k$. The transition model learns to predict the representations at the next time-step: $\pred\ttwo^k = \slot\tone^k + \transition([\slot\t^k, \slot\tone^k, \slot\tone^k -  \slot\t^k])$,
where $[\cdot]$ denotes concatenation of vectors. 
The transition function $\transition$ consists of three steps: a linear down-projection, LayerNorm \citep{ba2016layer} and a linear transformation.

\textbf{Set Encoder} \ 
We create a global image representation by applying a DeepSets \citep{zaheer2017deep} model on the slots and future slot predictions: $\globalrep_{\slot\t} = \mlp \left(\layernorm\left(\sum_{k=1}^K(\mlp(\slot\t^k))\right)\right)\in\mathbb{R}^{D}$.
Similarly, $\globalrep_{\pred\t} = \deepsets(\pred\t)$.
Thus, the loss that is applied on $\globalrep$ is applied to the aggregated \textit{set} of slots.

\textbf{Set Contrastive Loss} \ 
For the training of our model, we use an inner product between the vector representations as scoring function $g$ and the InfoNCE loss \citep{oord2018representation}: 
\begin{align}
    \loss^i &= - \mathbb{E}_{\XMat} \left[ \log \frac{g(\globalrep_{\pred\t}^i, \globalrep_{\slot\t}^i)}{
    \sum_{\globalrep_{\slot\t\prime}^j \in \batch} g(\globalrep_{\pred\t}^i, \globalrep_{\slot\t\prime}^j) + 
    \sum_{\globalrep_{\pred\t\prime}^j \in \batch} g(\globalrep_{\pred\t}^i, \globalrep_{\pred\t\prime}^j)} \right], \label{eq:loss}
\end{align}
with $g(\globalrep_{\pred\t}, \globalrep_{\slot\t}) = \exp ( \globalrep_{\pred\t}^T \cdot \globalrep_{\slot\t} / \tau)$, where $\tau$ is a temperature constant. $i,j$ refer to the samples within the dataset $\XMat$. 
The positive pair $(\globalrep_{\pred\t}^i, \globalrep_{\slot\t}^i)$ contains the globally aggregated predicted representation and the slot representation of sample $i$ at time-step $t$.
We use both the globally aggregated predicted representations and slot representations  $\globalrep_{\pred\t\prime}^j, \globalrep_{\slot\t\prime}^j$ taken from all time-steps of all sequences within the batch $\batch$ as negative samples. Since this loss operates on \textit{sets} of representations, we call it the Set Contrastive Loss (SetCon).
\section{Related Work}
\textbf{Object Discovery} \ 
Most unsupervised approaches to object discovery or object-centric representation learning focus on models trained with reconstruction losses in pixel space \citep{greff2016tagger,locatello2020object,eslami2016attend,greff2017neural,nash2017multi,van2018relational,kosiorek2018sequential,greff2019multi,burgess2019monet,engelcke2019genesis,stelzner2019faster,crawford2019spatially,jiang2019scalable,lin2020space, veerapaneni2020entity}.
Out of these approaches, Tagger \citep{greff2016tagger}, NEM \citep{greff2017neural}, R-NEM \citep{van2018relational}, IODINE \citep{greff2019multi}, MONET \citep{burgess2019monet}, GENESIS \citep{engelcke2019genesis}, and Slot Attention \citep{locatello2020object} are most closely related to our approach, as they use a set of generic embeddings in the form of slots together with a segmentation mask to represent objects, but different from our approach they rely on a decoder back into pixel space for training. Relying on a decoder together with a loss in pixel space for training can bias the learning process to depend heavily on object size and pose challenges in scenes containing complex textures.

\textbf{Contrastive Learning} \ 
Contrastive learning \citep{chopra2005learning,hadsell2006dimensionality,gutmann2010noise,mnih2013learning,mikolov2013distributed} has enjoyed increasing popularity in learning image representations \citep{dosovitskiy2014discriminative, oord2018representation,hjelm2018learning}, with recent methods achieving similar performance as supervised approaches on ImageNet \citep{he2020momentum, chen2020simple,chen2020big}.
A recent line of work \citep{kipf2019contrastive,racah2020slot,huang2020set,anonymous2021systematic}, starting with the C-SWM model \citep{kipf2019contrastive}, explores the use of contrastive objectives for object discovery. Similar to our model, these approaches use transformations from temporal data as learning signal for discovering individual objects, as common transformations on static images (such as rotation or flipping) are likely unsuitable for this task. Different from our SetCon model, these approaches apply a contrastive loss solely on the slot level, which does not establish communication between slots and hence can suffer from failure modes where objects are ignored or represented multiple times in multiple slots.
\section{Experiments}
\setlength{\tabcolsep}{1pt}
\definecolor{gray}{rgb}{0.3, 0.3, 0.3}
\definecolor{gray2}{rgb}{0.4, 0.4, 0.4}

\begin{table}
\centering
\caption{MSE and ARI scores ($\times \num{1e-2}$; mean $\pm$ standard error for 3 seeds) for unsupervised object discovery in multi-object datasets.
\footnotesize{*we omit ARI on the GridWorld dataset as the score is not well behaved on single pixel segmentation masks.}}
\label{tab:results}
\begin{tabularx}{0.99\linewidth}{l@{\hskip 10pt}l@{\hskip 10pt}l@{\hskip 15pt}rY@{\hskip 10pt}rY@{\hskip 10pt}rY@{\hskip 10pt}rY}
\toprule
                         & Loss & Encoder & \multicolumn{4}{l}{Future Prediction $\pred\t$} & \multicolumn{4}{l}{Slot Representation $\slot\t$} \\
                         & & & \multicolumn{2}{l}{MSE} & \multicolumn{2}{l}{ARI} & \multicolumn{2}{l}{MSE} & \multicolumn{2}{l}{ARI} \\
\midrule
    \parbox[t]{2mm}{\multirow{6}{*}{\rotatebox[origin=c]{90}{\textbf{GridWorld}}}} & \textcolor{gray}{Reconstruction} & \textcolor{gray}{FM-MLP}  &                         \textcolor{gray}{0.828} & \textcolor{gray}{$\pm$ 0.367} & \textcolor{gray}{*}  &  &             \textcolor{gray}{0.014} & \textcolor{gray}{$\pm$ 0.004}    & \textcolor{gray}{*} & \\
    &           & \textcolor{gray}{Slot Attention} &                \textcolor{gray}{3.104} & \textcolor{gray}{$\pm$ 1.056} & \textcolor{gray}{*}  &  &             \textcolor{gray}{0.450} & \textcolor{gray}{$\pm$ 0.383}    & \textcolor{gray}{*} & \\\arrayrulecolor{gray2}\cline{2-11}\\[-0.8em] \arrayrulecolor{black}
    & Slotwise  & FM-MLP &                              5.507 & $\pm$ 2.576 & * &  &             3.685 & $\pm$ 2.758   & * &\\
    &           & Slot Attention &                        5.123 & $\pm$ 2.338 & * &  &              4.199 & $\pm$ 3.079   & * & \\
    & \textbf{SetCon}    & FM-MLP &                                 3.139 & $\pm$ 0.127 & * &  &              0.525 & $\pm$ 0.134     & * &\\
    &           & \textbf{Slot Attention} &                          \textbf{1.214} & $\pm$ 0.245 & * &  &              \textbf{0.072} & $\pm$ 0.064 & * & \\

    \midrule
    \parbox[t]{2mm}{\multirow{6}{*}{\rotatebox[origin=c]{90}{\textbf{Bouncing Balls}}}} & \textcolor{gray}{Reconstruction} & \textcolor{gray}{FM-MLP} &  \textcolor{gray}{2.708} & \textcolor{gray}{$\pm$ 1.073} &  \textcolor{gray}{68.4} & \textcolor{gray}{$\pm$ 15.0} & \textcolor{gray}{0.784} & \textcolor{gray}{$\pm$ 0.495}    & \textcolor{gray}{66.6} & \textcolor{gray}{$\pm$ 14.5} \\
    &           & \textcolor{gray}{Slot Attention} &                  \textcolor{gray}{4.077} & \textcolor{gray}{$\pm$ 0.663} &  \textcolor{gray}{92.8} &  \textcolor{gray}{$\pm$ 2.2} &            \textcolor{gray}{0.142} & \textcolor{gray}{$\pm$ 0.042}    & \textcolor{gray}{85.2} &  \textcolor{gray}{$\pm$ 0.5} \\\arrayrulecolor{gray2}\cline{2-11}\\[-0.8em] \arrayrulecolor{black}
    & Slotwise  & FM-MLP &                               8.229 & $\pm$ 1.128 &  39.0 & $\pm$ 10.0 &            4.518 & $\pm$ 0.574  & 35.2 &  $\pm$ 5.1\\
    &           & Slot Attention &                        7.394 & $\pm$ 1.241 &  14.3 & $\pm$ 12.2 &            2.866 & $\pm$ 0.639   & 16.8 &  $\pm$ 9.5 \\
    & \textbf{SetCon}    & FM-MLP &                                 9.334 & $\pm$ 0.891 &  49.6 & $\pm$ 10.5 &            5.331 & $\pm$ 0.472     & 38.7 & $\pm$ 11.1\\
    &           & \textbf{Slot Attention} &                          \textbf{5.541} & $\pm$ 0.811 &  \textbf{86.8} &  $\pm$ 2.2 &            \textbf{0.807} & $\pm$ 0.033 & \textbf{75.3} &  $\pm$ 3.8 \\

\bottomrule
\end{tabularx}
\end{table}

In this section, we first describe the datasets we use in our experiments, outline our evaluation methods and describe our results. Additional details about our experiments can be found in \cref{sec:details}.

\textbf{Datasets} \ 
We evaluate our method on two synthetic video datasets: Multi-Object GridWorld and Bouncing Balls. Both datasets describe the movement of three visually distinct objects on a two-dimensional, black background, and reuse the same three objects throughout all sequences. In the \textbf{Multi-Object GridWorld}, colored pixels represent objects and are restricted to move in four directions: up, down, left and right. Pixels pass through each other, overlapping one another in random order. In the \textbf{Bouncing Balls} dataset, colored balls move in continuous space, bouncing against one another. In both datasets, the objects are reflected by image boundaries. 

\textbf{Evaluation} \ 
We evaluate the performance of our models by training an IODINE spatial broadcast decoder \citep{greff2019multi,watters2019spatial} on top of the learned representations without propagating the gradients to the encoder and transition model. In line with previous work \citep{greff2019multi, locatello2020object}, this decoder decodes each slot individually into four channels, representing RGB colors and an unnormalized alpha mask. For the final output, the alpha masks are normalized across slots and used to combine the slotwise reconstructions. 
We apply this decoder both on the slot representations $\slot\t$ and the predictions $\pred\t$ created by the transition model, and measure two metrics: \textbf{1) Mean Squared Error (MSE)} between the reconstruction created by the decoder and the input $\inputt\t$ of the model. \textbf{2) Adjusted Rand Index (ARI)} \citep{rand1971objective,hubert1985comparing} between the normalized alpha masks of the decoder and the ground truth object segmentation (excl.~background). ARI measures clustering similarity, and ranges from 0 (random) to 1 (perfect match). 

\textbf{Baselines} \ 
We compare the SetCon Slot Attention model against several model variants, where we combine three different losses with two different encoder architectures. Next to the SetCon loss, we apply a slotwise contrastive loss similar to previous work \citep{kipf2019contrastive, racah2020slot, huang2020set, anonymous2021systematic}. For this loss, we replace the set representations $\globalrep_{\slot\t}, \globalrep_{\pred\t}$ in \cref{eq:loss} with the slot and transition model representations $\slot\t, \pred\t$, and contrast against the respective representations across samples and time-steps in the batch, so that each slot is paired with the slot of the same index in a different time-step or sample. Additionally, we compare these two losses against the ``ground-truth'' approach of using the reconstruction error of the decoder for training the entire model.
Next to the Slot Attention model \citep{locatello2020object}, we also test a C-SWM-like model \citep{kipf2019contrastive} for creating object-centric representations, which we coin FeatureMap MLP (FM-MLP). After the encoding backbone, it creates object representations by applying a linear layer with dimensionality corresponding to the number of objects and treating these features as slots.

\begin{figure}
    \centering
    \subfloat[\label{fig:examples}]{
        \includegraphics[height=10.5em]{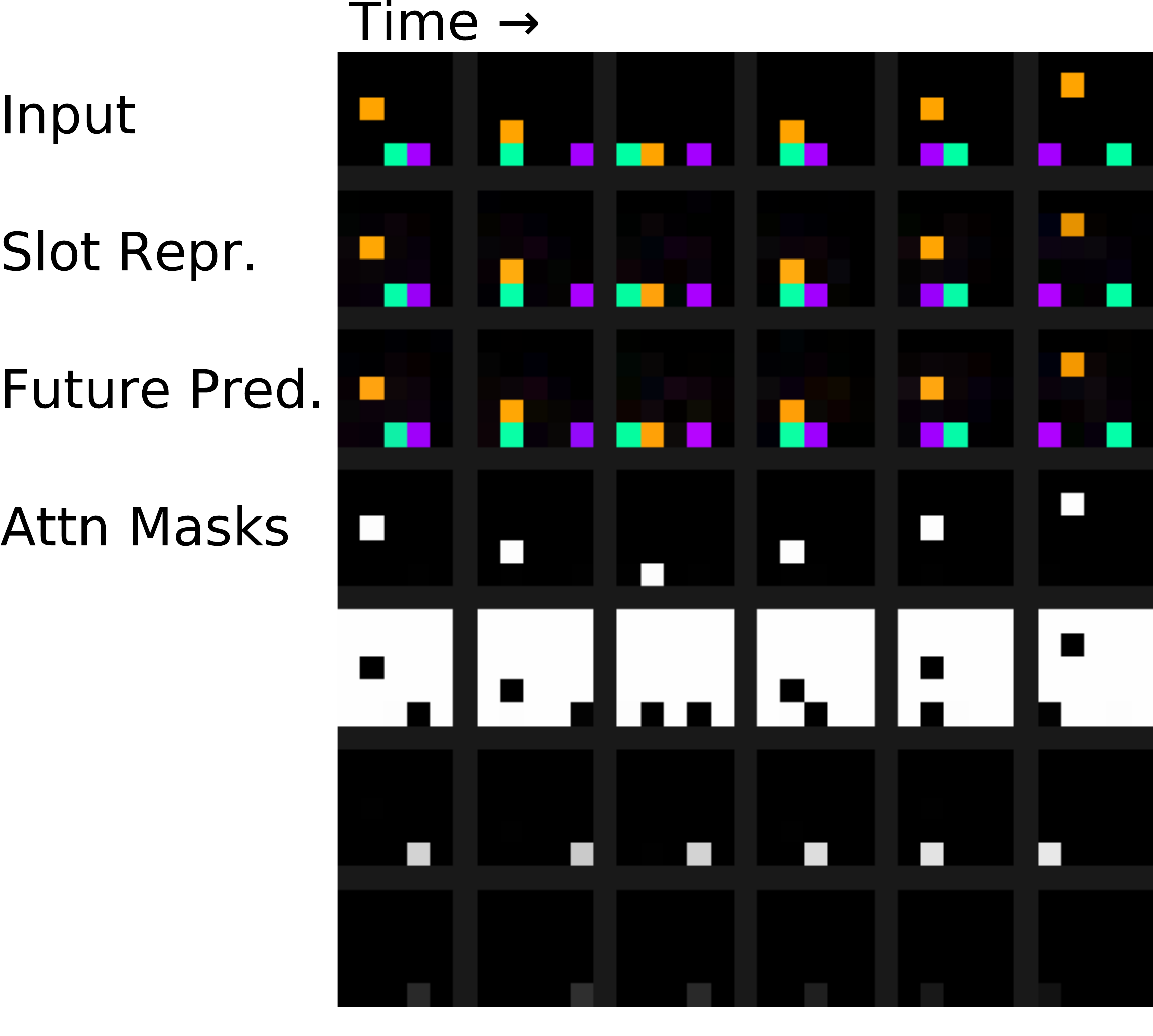} 
        \includegraphics[height=10.5em]{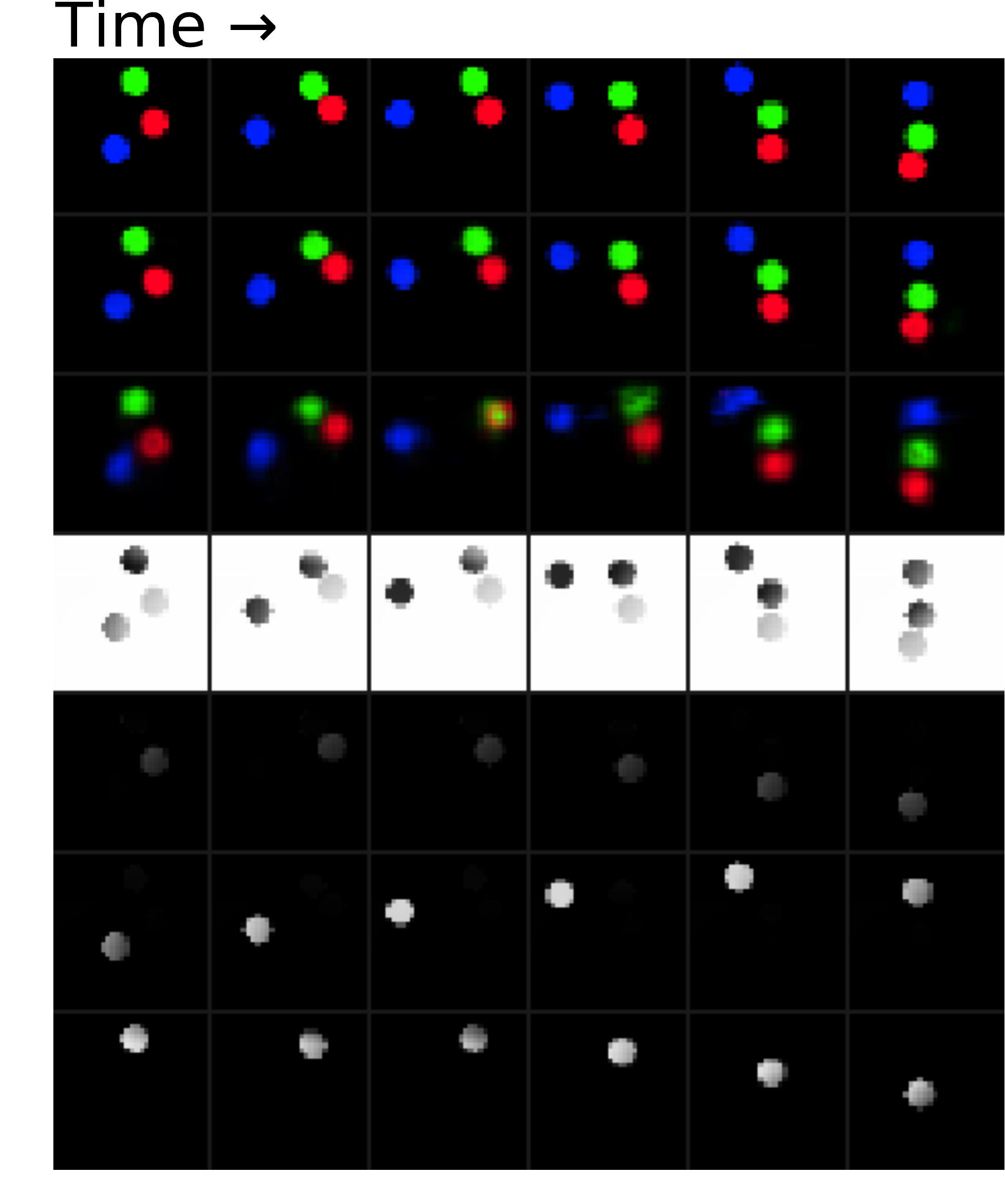}
        } %
        \,\,\,
    \subfloat[\label{fig:future_prediction_bb}]{
        \includegraphics[height=10.5em]{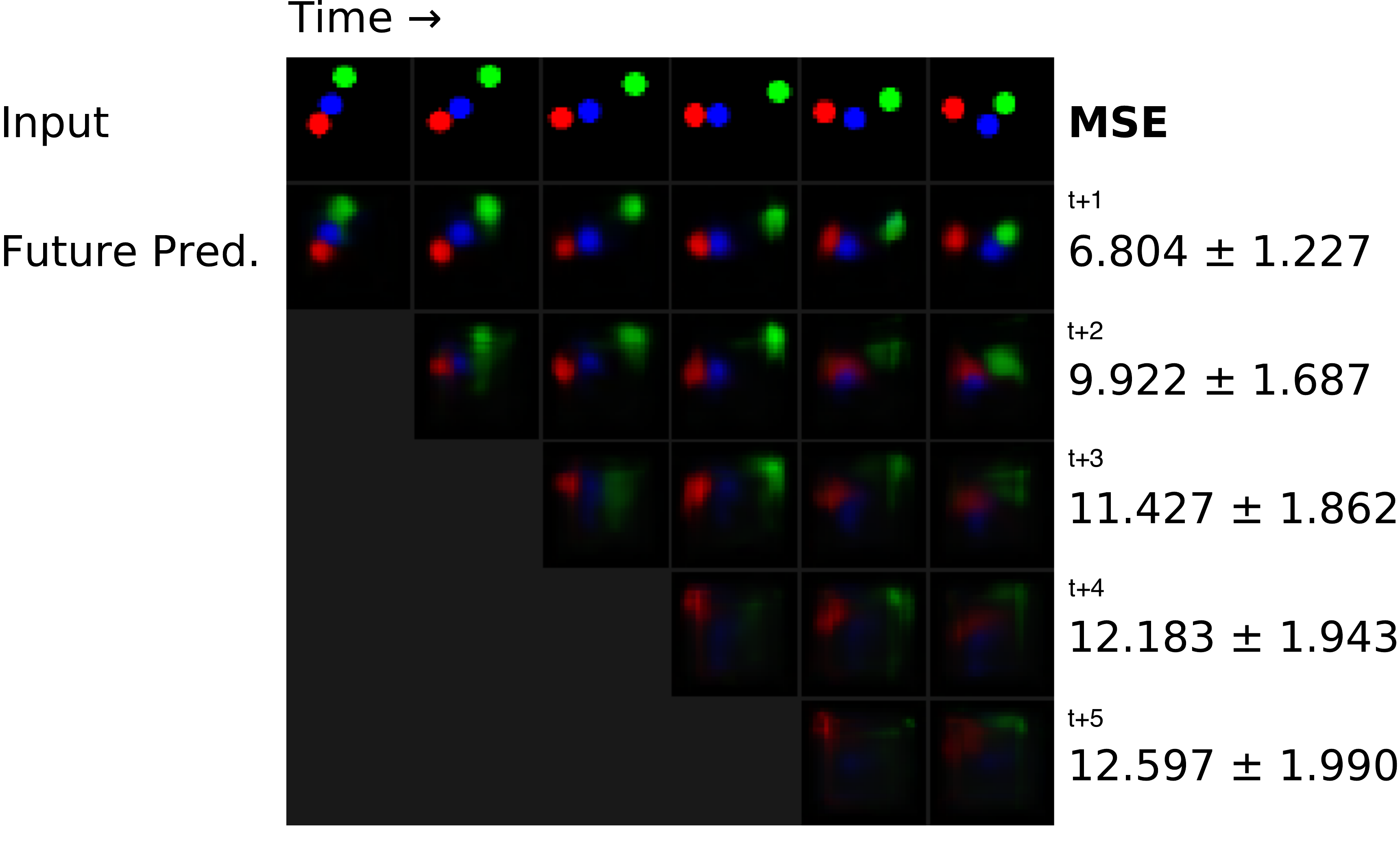}
    }
    \caption{Qualitative Results. 
    \textbf{(a)} Object separation and reconstruction quality for the SetCon Slot Attention model. \textbf{(b)} Reconstruction performance (MSE) when predicting up to five time-steps into the future (from top to bottom) on bouncing balls.}
    \vspace{-0.8em}
\end{figure}

\textbf{Results} \ 
As shown in \cref{tab:results}, SetCon with Slot Attention outperforms all other contrastive approaches across all metrics, and approaches the performance of the reference models trained with the reconstruction loss. For the FM-MLP model, we do not observe a significant performance difference between the SetCon and the slotwise contrastive loss on the Bouncing Balls dataset. In all other settings (FM-MLP model on the GridWorld, Slot Attention on both datasets), the SetCon loss consistently outperforms the slotwise contrastive loss, indicating that it does indeed enforce better coordination between slots and thus an improved object-centric representation. 
\cref{fig:examples} shows qualitative results of SetCon with Slot Attention. It becomes apparent from the attention masks (\textit{Attn Masks}) that the model learns to attend to different objects within the different slots\footnote{Note that on the GridWorld data the model seems to represent the third object within the ``background'' slot.}. 

\textbf{Future Prediction} \  \label{sec:future_pred}
In \cref{fig:future_prediction_bb}, we investigate whether our model can predict up to 5 time-steps into the future by applying the transition model iteratively without encoding a new input frame. We find that prediction quality can degrade quickly, as the model has only been trained for single-step prediction and has no explicit constraint to enforce the slot and transition model representations $\slot\t, \pred\t$ to lie nearby in latent space. This is an interesting avenue for future work.
\section{Conclusion}
Our results indicate that global, set-based contrasting (SetCon) can be an effective alternative learning paradigm for object-centric video models. While our exploration focuses on extremely simplistic environments and merely scratches the surface of this class of models, we are hopeful that further exploration will resolve current limitations for predicting further into the future, for disambiguating object instances with the same appearance, and will ultimately facilitate effective modeling of visually rich and diverse scenes, without having to rely on pixel-based losses.

\begin{ack}
We would like to thank Jean-Baptiste Cordonnier, Francesco Locatello,  Aravindh Mahendran, Thomas Unterthiner, Jakob Uszkoreit and Dirk Weissenborn for helpful discussions, and Marvin Ritter and Andreas Steiner for technical support. 
\end{ack}

\bibliographystyle{unsrtnat}
\bibliography{main}

\clearpage

\appendix
\section{Implementation Details}
\label{sec:details}

We implemented all our experiments in Jax \citep{jax2020github} and Flax \citep{flax2020github}.

\subsection{Model}
Unless specified differently, all $\mlp$ within our model refer to a two-layer MLP with hidden size of 128 and ReLU activation. Each $\mlp$ in our model has a separate set of trainable parameters.

\paragraph{Encoder} 
We implement the encoder $\backbone$ as follows:
\begin{align}
    \latentrep\t^{(1)} &= \mlp(\inputt\t) + \dense(\posemb) \\
    \latentrep\t &= \mlp(\layernorm(\latentrep\t^{(1)}))
\end{align}
$\mlp$ consists of two convolutional layers with kernelsize 1x1 with ReLU activation and filter-size $\filtersize=32$. $\dense$ is a single convolutional layers with kernelsize 1x1 and filter-size $\filtersize=32$.
$\posemb$ is a constant array that goes from from 0 to 1 in every image dimension.

\paragraph{Slot Attention}
For Slot Attention, we followed the implementation provided by the authors\footnote{\href{https://github.com/google-research/google-research/tree/master/slot\_attention}{https://github.com/google-research/google-research/tree/master/slot\_attention}}. The feedforward network $\ffw$ consists of a GRU~\citep{cho2014learning} and an MLP with skip connections. Following the original implementation of Slot Attention, we use LayerNorm on input features $\latentrep\t$ and initial slot representations $\slotinit$.
Unless specified differently, we use four slots of dimensionality $\slotsize=16$. As we are only considering simplistic environments where each object is assigned a unique, fixed color, we do not need to break symmetries between objects of identical appearance. Thus, we run Slot Attention with slot-specific learnable initializations and only a single iteration of the attention mechanism. We leave using a symmetric treatment of slots with multiple iterations of Slot Attention for future work.

\paragraph{FeatureMap MLP}
The FeatureMap MLP (FM-MLP) makes use of the same encoder but creates slot representations by applying a linear layer with each feature being interpreted as an individual slot. Each slot is then further processed through an $\mlp$:
\begin{align}
    \slot\t &= \reshape(\dense(\latentrep\t)) \\
    \slot\t &= \mlp(\slot\t) \label{eq:featuremap_mlp}
\end{align}
In our experiments, since we want to model 3 objects and the background, we use four slots and thus $\dense$ has a filter-size of four.

\paragraph{Decoder}
We use a filter-size of 16 within the decoder on the GridWorld dataset and 32 on the Bouncing Balls dataset. The decoder is trained using the same learning rate as is used for the rest of the model.

\subsection{Training}
We train all our models for 100,000 steps using Adam \citep{kingma2014adam} and a weight decay of $\num{1e-6}$. We set the temperature $\tau$ in the score function of the SetCon loss to 0.5.
We use a constant learning rate $lr$, scale it depending on the batch-size $b$: $lr \cdot \frac{b}{256}$, and tune it:
We ran all models with 5 learning rates $lr = (\num{1e-4}, \num{2e-4}, \num{3e-4}, \num{4e-4}, \num{5e-4})$ and selected the best result based on their MSE. On the bouncing ball dataset, we select the best model based on its performance on the training set and report the results on a separate test-set. 

We trained all our models on 8 TPU v2 cores.

In the GridWorld dataset, we use a batch-size of 1024 which is divided equally between the 8 TPU v2 cores that we train on. In the Bouncing Balls dataset, we set the global batch-size to 128.

\subsection{Datasets}

\paragraph{Multi-Object GridWorld}
The Multi-Object GridWorld dataset consists of sequences of eight frames, where each frame is made up of a 5x5 black plain. Colored pixels representing objects move deterministically through the scene: each time-step they move exactly one pixel-length to the top, bottom, right or left, and at the image boundary they bounce off in the opposite direction. The starting position, direction and color of each pixel is randomly sampled for each sequence. Objects start at different locations, but might overlap throughout the time-series in which case one of them becomes invisible. 
Unless otherwise specified, we restrict the number of pixels (i.e. objects) and the set of possible colors to three, and ensure that each color appears exactly once per sequence. We generate this set of colors by selecting equally spaced out hue values in HSV with a random offset and converting them to RGB. 

Sequences are generated on the fly throughout training and evaluation. With 25 possible starting positions and 4 possible directions, this gives us a total of 970,200 possible sequences to generate. We evaluate all methods across the 5,000 last training batches, evaluating the performance on each batch before using its result for optimizing the model parameters.

\paragraph{Bouncing Balls}
The Bouncing Balls dataset consists of 2D sequences with three balls bouncing against one another and the image boundary. We generate 1,000 sequences and use 128 of these for evaluation. Each sequence consists of twelve frames. We generate three colors in the same way as for the Multi-Object GridWorld dataset, and use these across all sequences. Even though the training set is relatively small, we do not observe any overfitting in any of our models.

Note that due to the color selection in both datasets, it is possible to identify the objects by simply clustering the image by color.

We do not apply any random augmentations on both datasets, and solely rely on temporal differences for the contrastive loss. We scale all input values to the interval [-1, 1].

\subsection{Evaluation}
\paragraph{ARI Score} In line with previous works \citep{greff2019multi,burgess2019monet,engelcke2019genesis,locatello2020object}, we exclude background labels from the ARI evaluation. We use the implementation provided by \citet{multiobjectdatasets19}, available at \href{https://github.com/deepmind/multi_object_datasets}{https://github.com/deepmind/multi\_object\_datasets}.

\paragraph{Future Prediction} We train the decoder to decode all $\pred_{t+k}$ and $\slot\t$ to the respective inputs $\inputt_{t+k}$ and $\inputt\t$, and measure their reconstruction performance in MSE. We predict up to $k=5$ steps into the future. 

\section{Additional Experiments}

\subsection{Additional Results}

\begin{figure}
    \centering
    \includegraphics[height=12em]{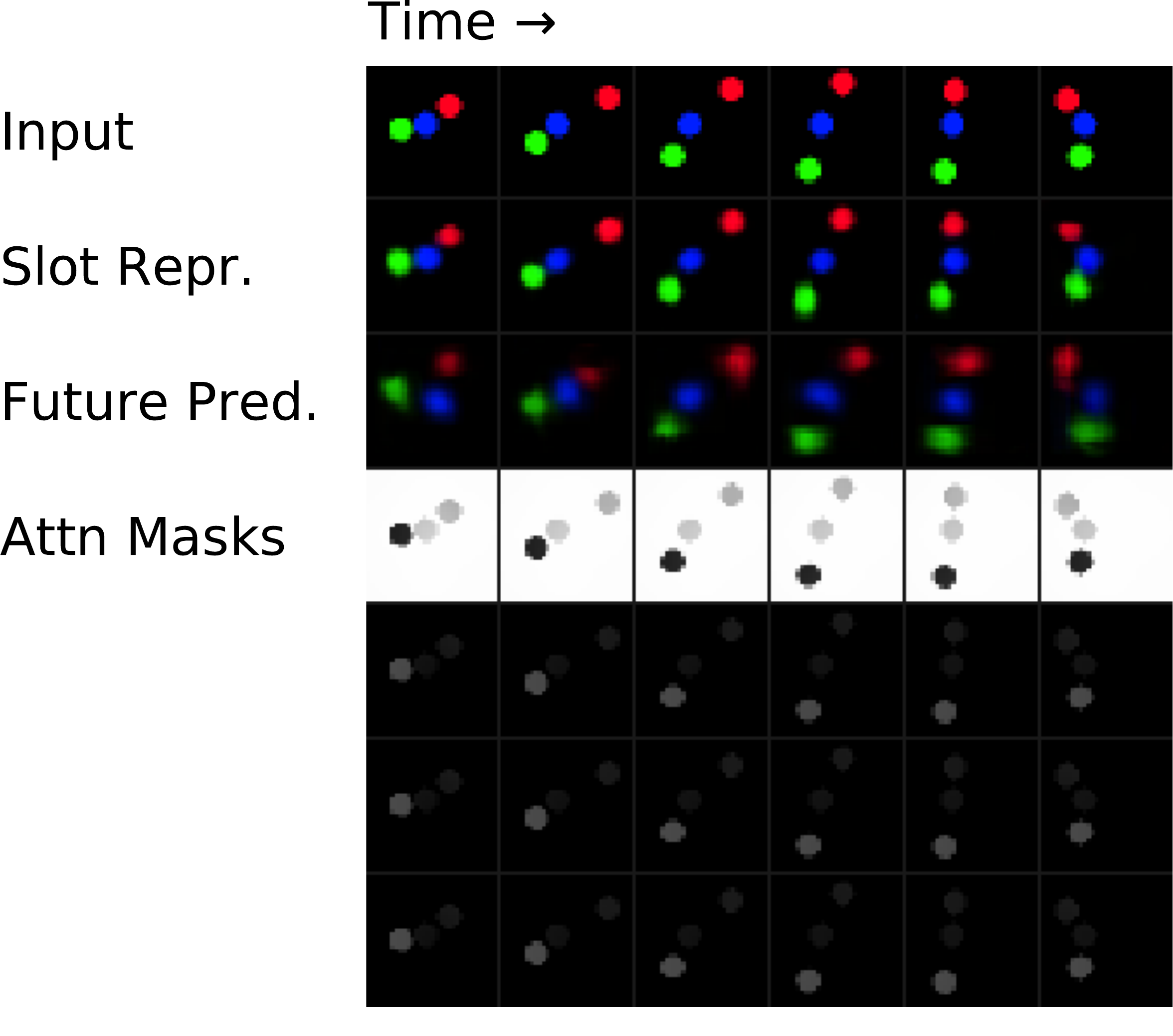}
    \includegraphics[height=12em]{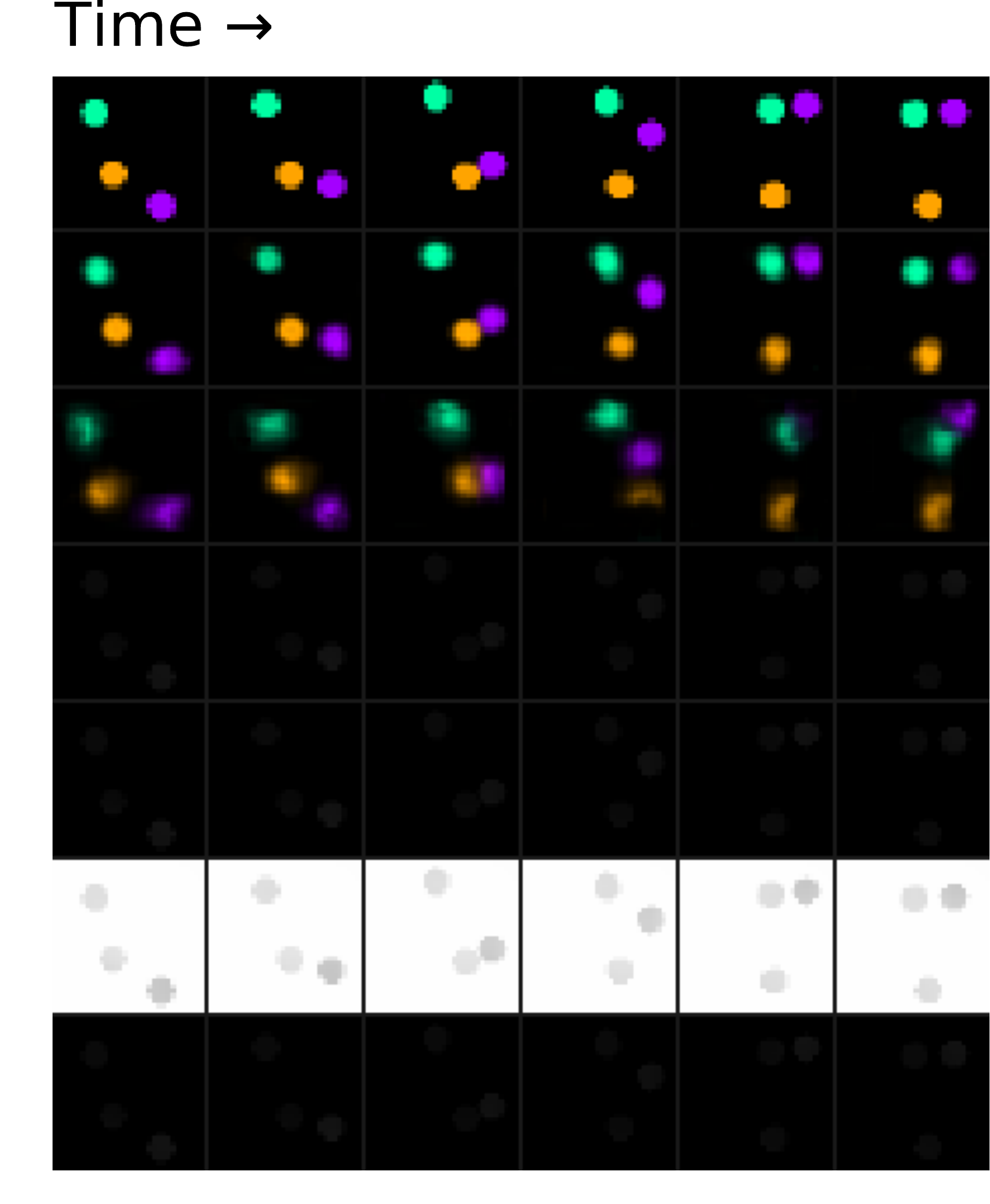}
    \caption{Qualitative Results for the Slot Attention model trained with the slotwise contrastive loss on the Bouncing Balls dataset.}
    \label{fig:slotwise}
\end{figure}

From the qualitative results of the Slot Attention model when trained with the slotwise contrastive loss on the Bouncing Balls dataset in \cref{fig:slotwise} it becomes apparent that this objective does not manage to push the individual slots to create diverse representations. 

While we tested only one setup of a slotwise loss in our experiments, we expect the described fundamental problem to hold for other versions as well.
In our experiments, we tested a slotwise loss as presented by \citet{kipf2019contrastive} in which slots are paired based on their index. %
Alternatively, one could design a slotwise loss in which slots with different indices are matched based on their respective loss. We do not expect such a setup to resolve the failure case of slotwise losses presented in this paper. It does not enforce any communication between the slots within the loss and thus it would still be a feasible solution for the model to learn to represent the same object across all slots.

\begin{figure}
    \centering
    \includegraphics[height=12em]{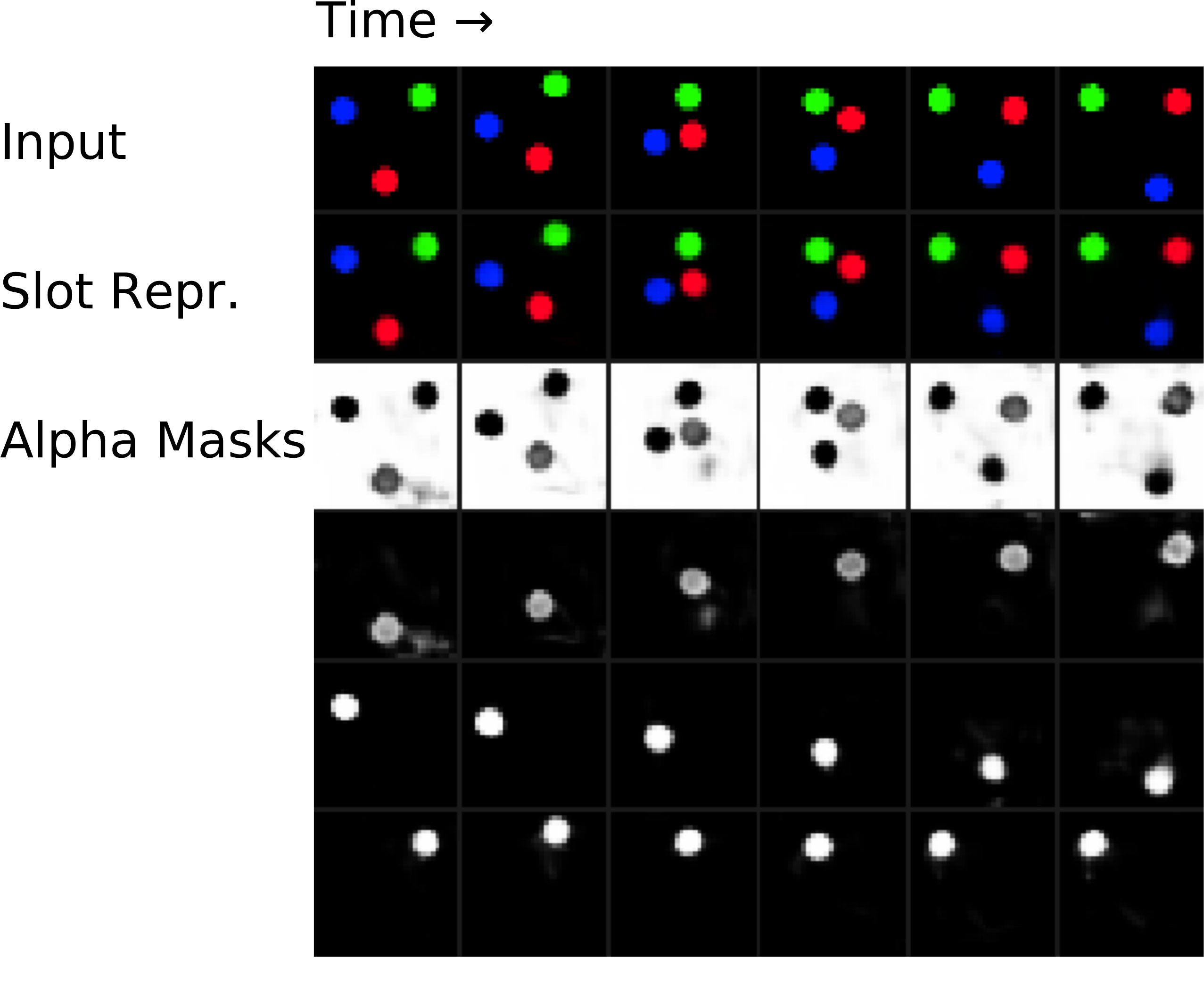}
    \includegraphics[height=12em]{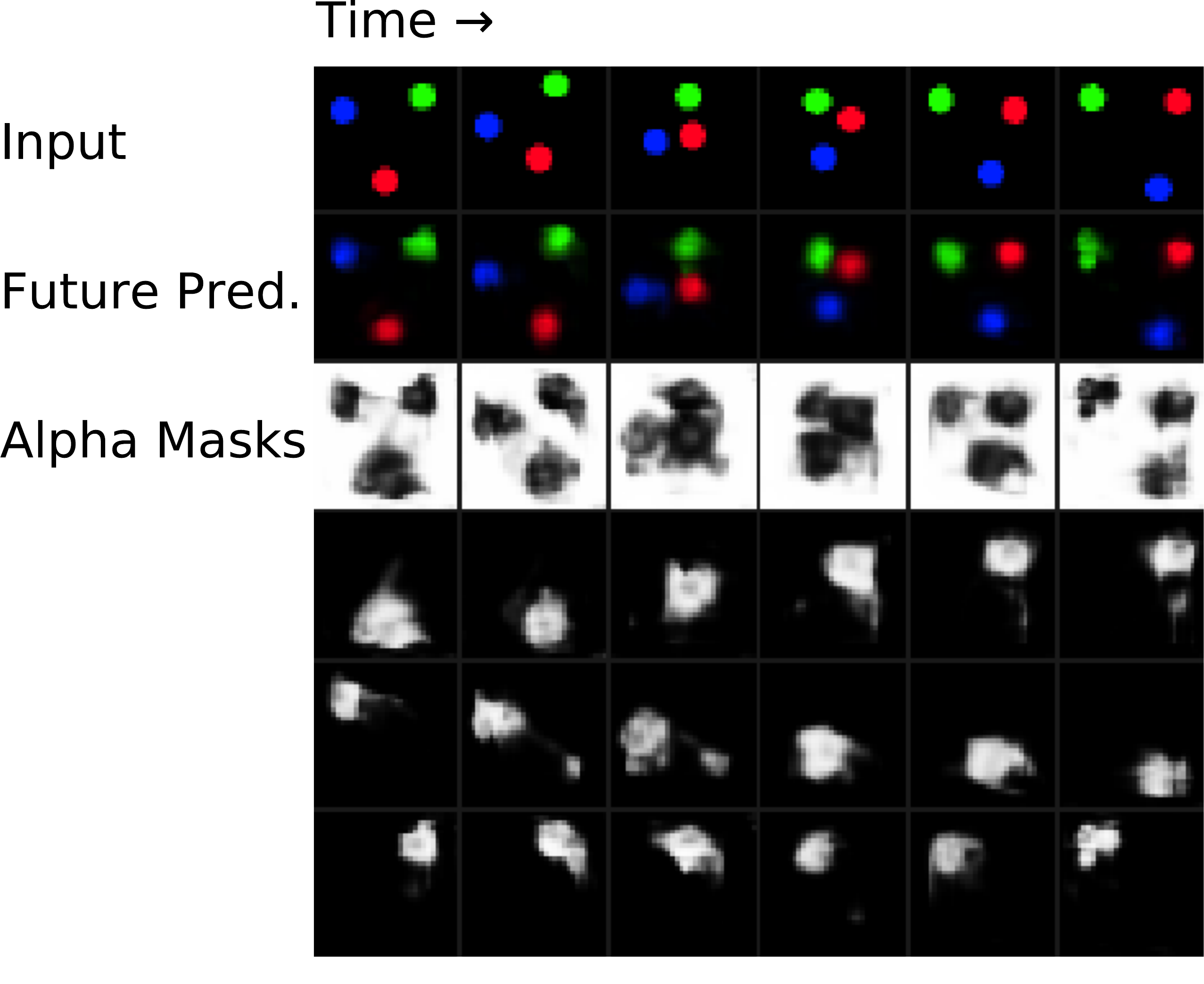}
    \caption{Decoder: Normalized alpha masks when trained on top of the SetCon Slot Attention model.}
    \label{fig:alpha_masks}
\end{figure}

\cref{fig:alpha_masks} shows the normalized alpha masks of the decoder for the SetCon Slot Attention model.

\subsection{Future Prediction}

\begin{figure}
    \centering
    \includegraphics[height=11em]{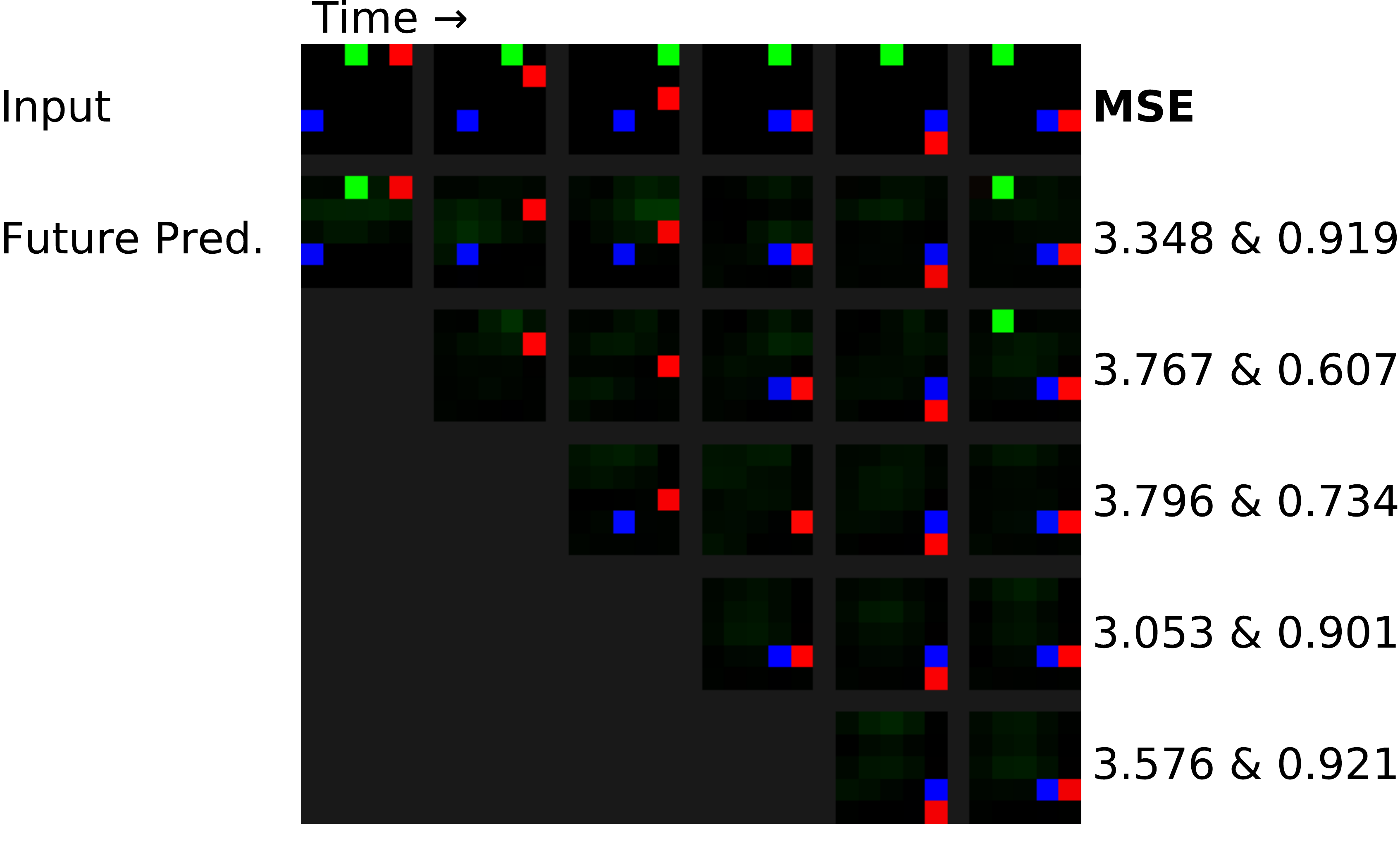}
    \caption{Predicting several time-steps into the future. From top to bottom: predicting further into the future.}
    \label{fig:future_prediction_gw}
\end{figure}

Similarly to \cref{sec:future_pred}, we test whether our model is capable of predicting 5 time-steps into the future on the GridWorld dataset.
As shown in \cref{fig:future_prediction_gw}, our approach leads to an overall worse future prediction performance compared to our main result (\cref{tab:results}). Nonetheless, it achieves a relatively consistent performance across all predicted time-steps, and still performs favorably compared to the single-step prediction of the slotwise contrastive loss. 
The qualitative example shows that the model does manage to accurately predict the future trajectory of some objects (here: red pixel), but largely fails for others (here: green pixel), explaining the overall worse, but consistent performance.

\subsection{Number of Colors}
\begin{figure}
    \centering
    \includegraphics[width=0.8\textwidth]{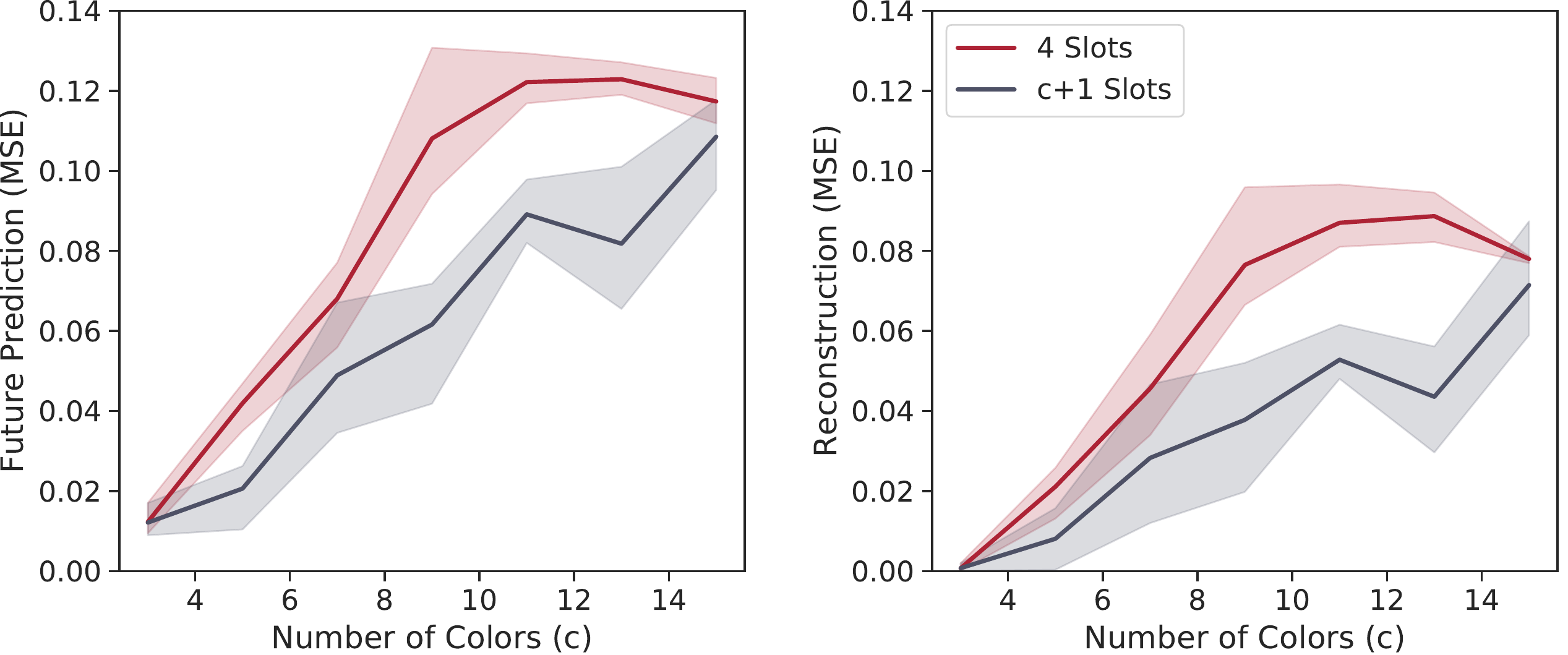}
    \caption{Reconstruction performance (MSE) depending on the number of colors and slots in the GridWorld dataset.
    }
    \label{fig:numcolors}
\end{figure}

In \cref{fig:numcolors}, we evaluate the performance of the SetCon Slot Attention model when adding more colors to the GridWorld dataset. The plots show that the performance of the model degrades when more colors are added, irrespective of whether we use a fixed number of slots (four), or whether we use one slot for each color (plus one for the background). This indicates that the SetCon model relies heavily on the color information to separate different objects, and to create a better representation of the scene and its dynamics.

\subsection{Random Initialization of Slots}
Due to the learned initialization of the slots within the Slot Attention model, our model looses the ability to generalize to scenes with more objects. In this experiment, we want to investigate how our approach performs under randomly initialized slots. For this, we randomly sample from a unit Gaussian distribution $\boldsymbol{\varepsilon} \sim \mathcal{N}(0,1)$ and scale the result using learned parameters $\boldsymbol{\mu}, \boldsymbol{\sigma}$ that are shared across slots: $\boldsymbol{c} = \boldsymbol{\mu} + \boldsymbol{\varepsilon} \cdot \boldsymbol{\sigma}$. Then, we use three iterations within the Slot Attention module to compute our final slot representations $\slot$.

We find that the performance deteriorates strongly with this approach: we achieve a MSE of $6.016 \pm 2.418$ on the future prediction $\pred\t$ and of $3.678 \pm 2.367$ on the slot representations $\slot\t$. This suggests that further exploration is necessary to make our approach work for scenes that require decomposition of objects with identical appearance (such as two red objects in one scene).

\end{document}